\documentclass[sigconf]{acmart}

\usepackage{array}
\usepackage{graphicx}
\usepackage{multirow}
\usepackage{amsmath}
\usepackage{amsfonts}
\usepackage{amssymb}
\usepackage{MnSymbol}
\usepackage[mathscr]{euscript}
\usepackage{tabularx}
\usepackage{float}

\newcolumntype{x}[1]{>{\centering\arraybackslash}p{#1}}
\newcolumntype{Y}{>{\centering\arraybackslash}X}
\newcommand{\trace}{\operatorname{tr}}

\newcommand{\vvec}{\operatorname{vec}}

\AtBeginDocument{%
  \providecommand\BibTeX{{%
    \normalfont B\kern-0.5em{\scshape i\kern-0.25em b}\kern-0.8em\TeX}}}

\copyrightyear{2020}
\acmYear{2020}
\setcopyright{acmcopyright}
\acmConference[KDD '20]{Proceedings of the 26th ACM SIGKDD Conference on Knowledge Discovery and Data Mining USB Stick}{August 23--27, 2020}{Virtual Event, USA}
\acmBooktitle{Proceedings of the 26th ACM SIGKDD Conference on Knowledge Discovery and Data Mining USB Stick (KDD '20), August 23--27, 2020, Virtual Event, USA}
\acmPrice{15.00}
\acmDOI{10.1145/3394486.3403065}
\acmISBN{978-1-4503-7998-4/20/08}

\begin{document}


\fancyhead{} 
\title{Kronecker Attention Networks}


\author{Hongyang Gao}
\affiliation{%
  \institution{Texas A\&M University}
  \city{College Station}
  \state{TX}
  \postcode{77843}
}
\email{hongyang.gao@tamu.edu}

\author{Zhengyang Wang}
\affiliation{%
  \institution{Texas A\&M University}
  \city{College Station}
  \state{TX}
  \postcode{77843}
}
\email{zhengyang.wang@tamu.edu}

\author{Shuiwang Ji}
\affiliation{%
  \institution{Texas A\&M University}
  \city{College Station}
  \state{TX}
  \postcode{77843}
}
\email{sji@tamu.edu}

\renewcommand{\shortauthors}{H. Gao, et al.}

\begin{abstract}
    Attention operators have been applied on both 1-D data like texts
    and higher-order data such as images and videos. Use of attention
    operators on high-order data requires flattening of the spatial or
    spatial-temporal dimensions into a vector, which is assumed to
    follow a multivariate normal distribution. This not only incurs
    excessive requirements on computational resources, but also fails to
    preserve structures in data. In this work, we propose to avoid
    flattening by assuming the data follow matrix-variate normal
    distributions. Based on this new view, we develop Kronecker
    attention operators (KAOs) that operate on high-order tensor data
    directly. More importantly, the proposed KAOs lead to dramatic
    reductions in computational resources. Experimental results show
    that our methods reduce the amount of required computational
    resources by a factor of hundreds, with larger factors for
    higher-dimensional and higher-order data. Results also show that
    networks with KAOs outperform models without attention, while
    achieving competitive performance as those with original attention
    operators.
\end{abstract}



\maketitle

\section{Introduction}

Deep neural networks with attention operators have shown great
capability of solving challenging tasks in various fields, such
as natural language
processing~\cite{bahdanau2014neural,vaswani2017attention,johnson2015semi},
computer vision~\cite{xu2015show,lu2016hierarchical}, and network
embedding~\cite{velivckovic2017graph,gao2018large}. Attention
operators are able to capture long-range dependencies, resulting
in significant performance
boost~\cite{li2018non,malinowski2018learning}. While attention
operators were originally proposed for 1-D data, recent
studies~\cite{wang2018non,zhao2018psanet,gao2019graph} have
attempted to apply them on high-order data, such as images and
videos. However, a practical challenge of using attention
operators on high-order data is the excessive requirement
computational resources, including computational cost and memory
usage. For example, for 2-D image tasks, the time and space
complexities are both quadratic to the product of the height and
width of the input feature maps. This bottleneck becomes
increasingly severe as the spatial or spatial-temporal dimensions
and the order of input data increase. Prior methods address this
problem by either down-sampling data before attention
operators~\cite{wang2018non} or limiting the path of
attention~\cite{huang2018ccnet}.

In this work, we propose novel and efficient attention operators,
known as Kronecker attention operators~(KAOs), for high-order data.
We investigate the above problem from a probabilistic perspective.
Specifically, regular attention operators flatten the data and
assume the flattened data follow multivariate normal distributions.
This assumption not only results in high computational cost and
memory usage, but also fails to preserve the spatial or
spatial-temporal structures of data. We instead propose to use
matrix-variate normal distributions to model the data, where the
Kronecker covariance structure is able to capture relationships
among spatial or spatial-temporal dimensions. Based on this new
view, we propose our KAOs, which avoid flattening and operate on
high-order data directly. Experimental results show that KAOs are as
effective as original attention operators, while dramatically
reducing the amount of required computational resources. In
particular, we employ KAOs to design a family of efficient modules,
leading to our compact deep models known as Kronecker attention
networks~(KANets). KANets significantly outperform prior compact
models on the image classification task, with fewer parameters and
less computational cost. Additionally, we perform experiments on
image segmentation tasks to demonstrate the effectiveness of our
methods in general application scenarios.

\section{Background and Related Work}

\begin{figure}[t] \includegraphics[width=\columnwidth]{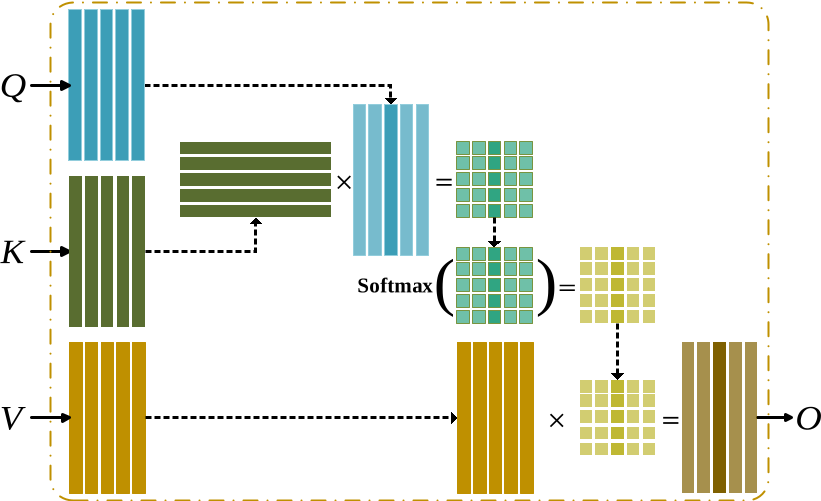}
    \Description[]{}
    \caption{An illustration of the attention operator.
        Here, $\times$ denotes matrix multiplication, and softmax($\cdot$)
        is the column-wise softmax operator. $\boldsymbol Q$, $\boldsymbol K$,
        and $\boldsymbol V$ are input matrices. A similarity score is
        computed between each query vector as a column of $\boldsymbol Q$
        and each key vector as a column in $\boldsymbol K$. Softmax($\cdot$)
        normalizes these scores and makes them sum to 1.
        Multiplication between normalized scores and the matrix $\boldsymbol
            V$ yields the corresponding output vector.}\label{fig:att_block}
\end{figure}

In this section, we describe the attention and related non-local
operators, which have been applied on various types of data such as
texts, images and videos.

\subsection{Attention Operator}\label{sec:att}

The inputs to an attention operator include a query matrix
$\boldsymbol Q = [\mathbf{q}_{1}, \mathbf{q}_{2}, \cdots,
    \mathbf{q}_{m}] \in \mathbb{R}^{d \times m}$ with each
$\mathbf{q}_{i} \in \mathbb{R}^{d}$, a key matrix $\boldsymbol K =
    [\mathbf{k}_{1}, \mathbf{k}_{2}, \cdots, \mathbf{k}_{n}] \in
    \mathbb{R}^{d \times n}$ with each $\mathbf{k}_{i} \in
    \mathbb{R}^{d}$, and a value matrix $\boldsymbol V =
    [\mathbf{v}_{1}, \mathbf{v}_{2}, \cdots, \mathbf{v}_{n}] \in
    \mathbb{R}^{p \times n}$ with each $\mathbf{v}_{i} \in
    \mathbb{R}^{p}$. The attention operation computes the responses of a
query vector $\mathbf{q}_i$ by attending it to all key vectors in
$\boldsymbol K$ and uses the results to take a weighted sum over
value vectors in $\boldsymbol V$. The layer-wise forward-propagation
operation of an attention operator can be expressed as
\begin{equation}\label{eq:att}
    \boldsymbol O = \mbox{attn}(\boldsymbol Q, \boldsymbol K,
    \boldsymbol V)=\boldsymbol V \times \mbox{Softmax}(\boldsymbol K^T
    \boldsymbol Q).
\end{equation}
Matrix multiplication between $\boldsymbol K^T$ and $\boldsymbol Q$
results in a coefficient matrix $\boldsymbol E=\boldsymbol K^T
    \boldsymbol Q$, in which each element $e_{ij}$ is calculated by the
inner product between $\mathbf{k}_{i}^T$ and $\mathbf{q}_{j}$.
This coefficient matrix $\boldsymbol E$ computes similarity scores between
every query vector $\mathbf{q}_i$, and every key vector
$\mathbf{k}_j$ and is normalized by a column-wise softmax operator
to make every column sum to 1.
The output $\boldsymbol O \in \mathbb{R}^{p \times m}$ is
obtained by multiplying $\boldsymbol V$ with the normalized
$\boldsymbol E$. In self-attention operators~\cite{vaswani2017attention},
we have $\boldsymbol Q=\boldsymbol K=\boldsymbol V$.
Figure~\ref{fig:att_block} provides an illustration of the attention
operator. The computational cost in Eq.~\ref{eq:att} is $O( m \times
    n \times (d+p))$. The memory required for storing the intermediate
coefficient matrix $\boldsymbol E$ is $O(mn)$. If $d = p$ and $m = n$,
the time and space complexities become $O(m^2 \times d)$ and $O(m^2)$,
respectively.

There are several other ways to compute $\boldsymbol E$ from $\boldsymbol Q$ and $\boldsymbol K$,
including Gaussian function, dot product, concatenation, and
embedded Gaussian function. It has been shown that dot product is
the simplest but most effective one~\cite{wang2018non}. Therefore,
we focus on the dot product similarity function in this work.

In practice, we can first perform separate linear transformations on each input
matrix, resulting in the following attention operator:
$\boldsymbol O = \boldsymbol W^V \boldsymbol V \mbox{Softmax}((\boldsymbol W^K \boldsymbol K)^T \boldsymbol W^Q \boldsymbol Q)$,
where $\boldsymbol W^V \in \mathbb{R}^{p' \times p}$, $\boldsymbol
    W^K \in \mathbb{R}^{d' \times d}$, and $\boldsymbol W^Q \in
    \mathbb{R}^{d' \times d}$. For notational simplicity, we omit linear
transformations in the following discussion.

\subsection{Non-Local Operator}\label{sec:nonlocal}

\begin{figure}[t]
    \Description[]{}
    \centering
    \includegraphics[width=\columnwidth]{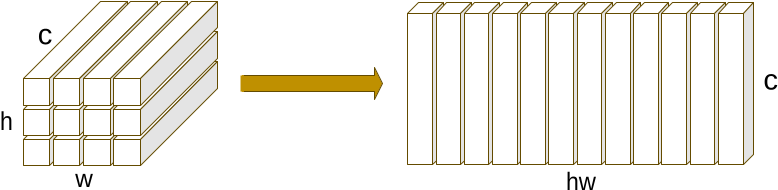}
    \caption{Conversion of a third-order tensor into a matrix by
        unfolding along mode-3. In this example, a $h \times w \times c$
        tensor is unfolded into a $c \times hw$ matrix.}\label{fig:unfold}
\end{figure}

Non-local operators, which is proposed in \cite{wang2018non}, apply
self-attention operators on higher-order data such as images and
videos. Taking 2-D data as an example, the input to the non-local operator
is a third-order tensor $\boldsymbol{\mathscr{X}} \in \mathbb{R}^{h\times w \times c}$,
where $h$, $w$, and $c$ denote the height, width, and number of
channels, respectively. The tensor is first converted into a matrix
$\boldsymbol X_{(3)} \in \mathbb{R}^{c \times hw}$ by unfolding
along mode-3~\cite{Tensor:Tamara}, as illustrated in Figure~\ref{fig:unfold}.
Then we perform the operation in Eq.~\ref{eq:att} by setting
$\boldsymbol Q = \boldsymbol K = \boldsymbol V = \boldsymbol X_{(3)}$.
The output of the attention operator is converted back to a third-order tensor as the
final output.

One practical challenge of the non-local operator is that it
consumes excessive computational resources. If $h = w$, the
computational cost of a 2-D non-local operator is $O(h^4 \times c)$.
The memory used to store the intermediate coefficient matrix incurs
$O(h^4)$ space complexity. The time and space complexities are
prohibitively high for high-dimensional and high-order data.

\section{Kronecker Attention Networks}

In this section, we describe our proposed Kronecker attention
operators, which are efficient and effective attention operators on
high-order data. We also describe how to use these operators to
build Kronecker attention networks.

\subsection{From Multivariate to Matrix-Variate Distributions}

We analyze the problem of attention operators on high-order data and
propose solutions from a probabilistic perspective. To illustrate
the idea, we take the non-local operator on 2-D data in
Section~\ref{sec:nonlocal} as an example.
Formally, consider a self-attention operator with $\boldsymbol Q =
    \boldsymbol K =\boldsymbol V = \boldsymbol X_{(3)}$, where
$\boldsymbol X_{(3)} \in \mathbb{R}^{c \times hw}$ is the mode-3
unfolding of a third-order input tensor $\boldsymbol{\mathscr{X}}
    \in \mathbb{R}^{h \times w \times c}$, as illustrated in
Figure~\ref{fig:unfold}. The $i$th row of $\boldsymbol X_{(3)}$
corresponds to $\vvec(\boldsymbol X_{::i})^T \in \mathbb{R}^{1
        \times hw}$, where $\boldsymbol X_{::i} \in \mathbb{R}^{h \times w}$
denotes the $i$th frontal slice of
$\boldsymbol{\mathscr{X}}$~\cite{Tensor:Tamara}, and $\vvec(\cdot)$
denotes the vectorization of a matrix by concatenating its
columns~\cite{gupta2018matrix}.

The frontal slices $\boldsymbol X_{::1}, \boldsymbol X_{::2},\ldots,
    \boldsymbol X_{::c} \in \mathbb{R}^{h \times w}$ of
$\boldsymbol{\mathscr{X}}$ are usually known as $c$ feature maps. In
this view, the mode-3 unfolding is equivalent to the vectorization
of each feature map independently. It is worth noting that, in
addition to $\vvec(\cdot)$, any other operation that transforms each
feature map into a vector leads to the same output from the
non-local operator, as long as a corresponding reverse operation is
performed to fold the output into a tensor. This fact indicates that
unfolding of $\boldsymbol{\mathscr{X}}$ in local operators ignores
the structural information within each feature map, \emph{i.e.,} the
relationships among rows and columns. In addition, such unfolding
results in excessive requirements on computational resources, as
explained in Section~\ref{sec:nonlocal}.

In the following discussions, we focus on one feature map
$\boldsymbol X \in \{\boldsymbol X_{::1}, \boldsymbol
    X_{::2},\ldots, \boldsymbol X_{::c}\}$ by assuming feature maps are
conditionally independent of each other, given feature maps of
previous layers. This assumption is shared by many deep learning
techniques that process each feature map independently, including
the unfolding mentioned above, batch
normalization~\cite{ioffe2015batch}, instance
normalization~\cite{ulyanov2016instance}, and pooling
operations~\cite{lecun1998gradient}. To view the problem above from
a probabilistic
perspective~\cite{ioffe2015batch,ulyanov2016instance}, the unfolding
yields the assumption that $\vvec(\boldsymbol X)$ follows a
multivariate normal distribution as $\vvec(\boldsymbol X) \sim
    \mathcal{N}_{hw} (\boldsymbol \mu, \boldsymbol \Omega)$, where
$\boldsymbol \mu \in \mathbb{R}^{hw}$ and $\boldsymbol \Omega \in
    \mathbb{R}^{hw \times hw}$. Apparently, the multivariate normal
distribution does not model relationships among rows and columns in
$\boldsymbol X$. To address this limitation, we propose to model
$\boldsymbol X$ using a matrix-variate normal
distribution~\cite{gupta2018matrix}, defined as below.

\textbf{Definition 1.} A random matrix $\boldsymbol A \in
    \mathbb{R}^{m\times n}$ is said to follow a matrix-variate normal
distribution $\mathcal{MN}_{m \times n}(\boldsymbol M,\boldsymbol
    \Omega \otimes \boldsymbol \Psi)$ with mean matrix $\boldsymbol M
    \in \mathbb{R}^{m \times n}$ and covariance matrix $\boldsymbol
    \Omega \otimes \boldsymbol \Psi$, where $\boldsymbol \Omega \in
    \mathbb{R}^{m \times m} \succ 0$ and $\boldsymbol \Psi \in
    \mathbb{R}^{n \times n} \succ 0$, if $\vvec(\boldsymbol A^T) \sim
    \mathcal{N}_{mn}(\vvec(\boldsymbol M^T), \boldsymbol \Omega \otimes
    \boldsymbol \Psi)$. Here, $\otimes$ denotes the Kronecker
product~\cite{van2000ubiquitous,graham2018kronecker}.

The matrix-variate normal distribution has separate covariance
matrices for rows and columns. They interact through the Kronecker
product to produce the covariance matrix for the original
distribution. Specifically, for two elements $X_{ij}$ and $X_{i'j'}$
from different rows and columns in $\boldsymbol X$, the relationship
between $X_{ij}$ and $X_{i'j'}$ is modeled by the interactions
between the $i$th and $i'$th rows and the $j$th and $j'$th columns.
Therefore, the matrix-variate normal distribution is able to
incorporate relationships among rows and columns.

\subsection{The Proposed Mean and Covariance Structures}\label{sec:MCS}

In machine learning, Kalaitzis et
al.~\cite{kalaitzis2013bigraphical} proposed to use the Kronecker
sum to form covariance matrices, instead of the Kronecker
product. Based on the above observations and studies, we propose
to model $\boldsymbol X$ as $\boldsymbol X \sim \mathcal{MN}_{h
        \times w}(\boldsymbol M,\boldsymbol \Omega \oplus \boldsymbol
    \Psi)$, where $\boldsymbol M \in \mathbb{R}^{h \times w}$,
$\boldsymbol \Omega \in \mathbb{R}^{h \times h} \succ 0$,
$\boldsymbol \Psi \in \mathbb{R}^{w \times w} \succ 0$, $\oplus$
denotes the Kronecker sum~\cite{kalaitzis2013bigraphical},
defined as $\boldsymbol \Omega \oplus \boldsymbol \Psi =
    \boldsymbol \Omega \otimes \boldsymbol I_{[w]}+ \boldsymbol
    I_{[h]} \otimes \boldsymbol \Psi$, and $\boldsymbol I_{[n]}$
denotes an $n \times n$ identity matrix. Covariance matrices
following the Kronecker sum structure can still capture the
relationships among rows and
columns~\cite{kalaitzis2013bigraphical}. It also follows
from \cite{allen2010transposable,wang2019spatial} that
constraining the mean matrix $\boldsymbol M$ allows a more direct
modeling of the structural information within a feature map.
Following these studies, we assume $\boldsymbol X$ follows a
variant of the matrix-variate normal distribution as
\begin{equation}
    \label{eqn:mvn_1} \boldsymbol X \sim \mathcal{MN}_{h \times
        w}(\boldsymbol M, \boldsymbol \Omega \oplus \boldsymbol \Psi),
\end{equation}
where the mean matrix $\boldsymbol M \in \mathbb{R}^{h \times w}$ is
restricted to be the outer sum of two vectors, defined as
\begin{equation}
    \boldsymbol M = \boldsymbol \mu \diamondplus \boldsymbol \upsilon =
    \boldsymbol \mu \mathbf{1}^T_{[w]} + \mathbf{1}_{[h]} \boldsymbol
    \upsilon^T,
\end{equation}
where $\boldsymbol \mu \in \mathbb{R}^{h}$,
$\boldsymbol \upsilon \in \mathbb{R}^{w}$, and $\mathbf{1}_{[n]}$
denotes a vector of all ones of size $n$.

\begin{figure*}[t] \includegraphics[width=\textwidth]{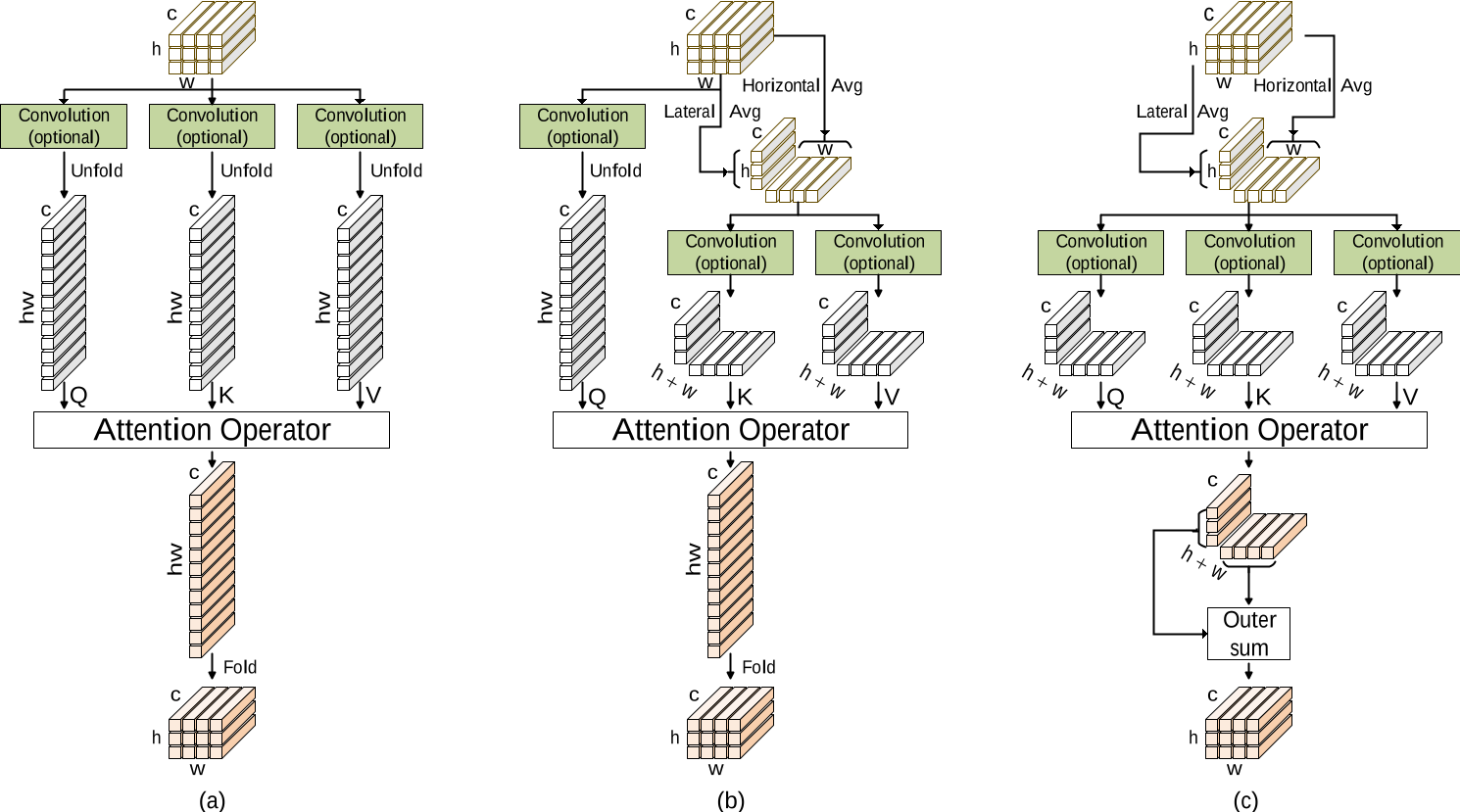}
    \Description[]{}
    \caption{Illustrations of regular attention operator~(a),
        KAO\textsubscript{{\tiny{$KV$}}}~(b) and KAO\textsubscript{{\tiny{$QKV$}}}~(c)
        on 2-D data. In the regular attention operator~(a), the input tensor is
        unfolded into a mode-3 matrix and fed into the attention operator. The output
        of the attention operator is folded back to a tensor as the final output. In
        KAO\textsubscript{{\tiny{$KV$}}}~(b), we juxtapose the horizontal and lateral average
        matrices derived from the input tensor as the key and value matrices. We keep
        the mode-3 unfolding of input tensor as the query matrix. In
        KAO\textsubscript{{\tiny{$QKV$}}}~(c), all three input matrices use the
        juxtaposition of two average matrices. In contrast to
        KAO\textsubscript{{\tiny{$KV$}}}, we use an outer-sum operation to generate the
        third-order tensor from the output of the attention operator.}\label{fig:atts}
\end{figure*}

Under this model, the marginal distributions of rows and columns are
both multivariate normal~\cite{allen2010transposable}. Specifically,
the $i$th row vector $\boldsymbol X_{i:} \in \mathbb{R}^{1 \times
        w}$ follows
$\boldsymbol X_{i:}^T \sim \mathcal{N}_{w} (\mu_i + \boldsymbol
    \upsilon^T, \Omega_{ii} + \boldsymbol \Psi)$,
and the $j$th column vector $\boldsymbol X_{:j} \in \mathbb{R}^{h
        \times 1}$ follows
$\boldsymbol X_{:j} \sim \mathcal{N}_{h} (\upsilon_i + \boldsymbol \mu, \Psi_{ii} + \boldsymbol \Omega)$.
In the following discussion, we assume that $\boldsymbol \Omega$ and
$\boldsymbol \Psi$ are diagonal, implying that any pair of variables
in $\boldsymbol X$ are uncorrelated. Note that, although the
variables in $\boldsymbol X$ are independent, their covariance
matrix still follows the Kronecker covariance structure, thus
capturing the relationships among rows and
columns~\cite{allen2010transposable,wang2019spatial}.

\subsection{Main Technical Results}\label{sec:tecresults}

Let $\overline{\boldsymbol X}_{row}= (\sum_{i=1}^{h}
    \boldsymbol X_{i:}^T)/h \in \mathbb{R}^{w}$ and $\overline{\boldsymbol
        X}_{col}=(\sum_{j=1}^{w}\boldsymbol X_{:j})/w \in
    \mathbb{R}^{h}$ be the average of row and column vectors,
respectively. Under the assumption above,
$\overline{\boldsymbol X}_{row}$ and $\overline{\boldsymbol
        X}_{col}$ follow multivariate normal distributions as
\begin{equation}
    \label{eqn:r_mean} \overline{\boldsymbol X}_{row} \sim
    \mathcal{N}_{w} (\overline{\boldsymbol \mu} + \boldsymbol \upsilon,
    \frac{\overline{\boldsymbol \Omega} + \boldsymbol \Psi}{h}),
\end{equation}
\begin{equation}
    \label{eqn:c_mean}
    \overline{\boldsymbol X}_{col} \sim \mathcal{N}_{h} (\overline{\boldsymbol \upsilon} + \boldsymbol \mu, \frac{\overline{\boldsymbol \Psi} + \boldsymbol \Omega}{w}),
\end{equation}
where $\overline{\boldsymbol \mu} = (\sum_{i=1}^{h}\mu_i)/h$,
$\overline{\boldsymbol \Omega} = (\sum_{i=1}^{h}\Omega_{ii})/h$,
$\overline{\boldsymbol \upsilon} = (\sum_{j=1}^{w}\upsilon_j)/w$,
and $\overline{\boldsymbol \Psi} = (\sum_{j=1}^{w}\Psi_{jj})/w$. Our
main technical results can be summarized in the following theorem.

\textbf{Theorem 1.} Given the multivariate normal distributions in
Eqs.~(\ref{eqn:r_mean}) and (\ref{eqn:c_mean}) with diagonal
$\boldsymbol \Omega$ and $\boldsymbol \Psi$, if \textit{(a)} $\mathbf{r}_1,
    \mathbf{r}_2, \ldots, \mathbf{r}_h$ are independent and identically
distributed~(i.i.d.) random vectors that follow the distribution in
Eq.~(\ref{eqn:r_mean}), \textit{(b)} $\mathbf{c}_1, \mathbf{c}_2, \ldots,
    \mathbf{c}_w$ are i.i.d. random vectors that follow the distribution
in Eq.~(\ref{eqn:c_mean}), \textit{(c)} $\mathbf{r}_1, \mathbf{r}_2, \ldots,
    \mathbf{r}_h$ and $\mathbf{c}_1, \mathbf{c}_2, \ldots, \mathbf{c}_w$
are independent, we have
\begin{equation}
    \label{eqn:mvn_2} \tilde{\boldsymbol X} \sim
    \mathcal{MN}_{h \times w}\left(\tilde{\boldsymbol M},
    \frac{\overline{\boldsymbol \Psi} + \boldsymbol \Omega}{w} \oplus
    \frac{\overline{\boldsymbol \Omega} + \boldsymbol \Psi}{h}\right),
\end{equation}
where $\tilde{\boldsymbol X} = [\mathbf{r}_1, \mathbf{r}_2, \ldots,
    \mathbf{r}_h]^T + [\mathbf{c}_1, \mathbf{c}_2, \ldots,
    \mathbf{c}_w]$, $\tilde{\boldsymbol M} = (\boldsymbol \mu
    \diamondplus  \boldsymbol \upsilon) + (\overline{\boldsymbol \mu} +
    \overline{\boldsymbol \upsilon})$. In particular, if $h = w$, the
covariance matrix satisfies
\begin{equation}
    \trace\left(\frac{\overline{\boldsymbol \Psi} + \boldsymbol
        \Omega}{w} \oplus \frac{\overline{\boldsymbol \Omega} + \boldsymbol
        \Psi}{h}\right) = \frac{2}{h} \trace\left(\boldsymbol \Omega \oplus
    \boldsymbol \Psi\right),
\end{equation}
where $\trace(\cdot)$ denotes matrix trace.

\begin{proof}
    The fact that $\boldsymbol \Omega$ and $\boldsymbol \Psi$ are
    diagonal implies independence in the case of multivariate normal
    distributions. Therefore, it follows from assumptions (a) and (b)
    that
    \begin{equation}
        \hspace{-0.2cm}[\mathbf{r}_1, \mathbf{r}_2, \ldots, \mathbf{r}_h]^T
        \sim \mathcal{MN}_{h \times w}\left( \boldsymbol M_r, \boldsymbol
        I_{[h]} \otimes \frac{\overline{\boldsymbol \Omega} + \boldsymbol
            \Psi}{h}\right),
    \end{equation}
    where $\boldsymbol M_r = \overline{\boldsymbol \mu} + [\boldsymbol \upsilon, \boldsymbol \upsilon, \ldots, \boldsymbol \upsilon]^T = \overline{\boldsymbol \mu} + \mathbf{1}_{[h]} \boldsymbol \upsilon^T$,
    and
    \begin{equation}
        \hspace{-0.2cm}[\mathbf{c}_1, \mathbf{c}_2, \ldots, \mathbf{c}_w]
        \sim \mathcal{MN}_{h \times w}\left( \boldsymbol M_c,
        \frac{\overline{\boldsymbol \Psi} + \boldsymbol \Omega}{w} \otimes
        \boldsymbol I_{[w]}\right),
    \end{equation}
    where $\boldsymbol M_c = \overline{\boldsymbol \upsilon} + [\boldsymbol \mu, \boldsymbol \mu, \ldots, \boldsymbol \mu] = \overline{\boldsymbol \upsilon} + \boldsymbol \mu \mathbf{1}^T_{[w]}.$

    Given assumption (c) and $\tilde{\boldsymbol X} = [\mathbf{r}_1, \mathbf{r}_2, \ldots, \mathbf{r}_h]^T + [\mathbf{c}_1, \mathbf{c}_2, \ldots, \mathbf{c}_w]$, we have
    \begin{equation}
        \tilde{\boldsymbol X} \sim \mathcal{MN}_{h \times
            w}\left(\tilde{\boldsymbol M}, \frac{\overline{\boldsymbol \Psi} +
            \boldsymbol \Omega}{w} \oplus \frac{\overline{\boldsymbol \Omega} +
            \boldsymbol \Psi}{h}\right),
    \end{equation}
    where $\tilde{\boldsymbol M} = \boldsymbol M_r + \boldsymbol M_c = (\boldsymbol \mu \diamondplus \boldsymbol \upsilon) + (\overline{\boldsymbol \mu} + \overline{\boldsymbol \upsilon})$.

    If $h=w$, we have
    \begin{equation}
        \trace(\boldsymbol \Omega \oplus \boldsymbol \Psi) = h\left(\sum
        \Omega_{ii} + \sum \Psi_{jj}\right),
    \end{equation}
    and
    \begin{eqnarray}
        &&\trace\left(\frac{\overline{\boldsymbol \Psi} + \boldsymbol \Omega}{w} \oplus \frac{\overline{\boldsymbol \Omega} + \boldsymbol \Psi}{h}\right) \nonumber \\
        &=& \trace\left(\frac{1}{h}(\boldsymbol \Omega \oplus \boldsymbol \Psi) + \frac{1}{h}(\overline{\boldsymbol \Psi} + \overline{\boldsymbol \Omega})\right) \nonumber \\
        &=& \left(\sum \Omega_{ii} + \sum \Psi_{jj}) + h(\overline{\boldsymbol \Psi} + \overline{\boldsymbol \Omega}\right) \nonumber \\
        &=& 2(\sum \Omega_{ii} + \sum \Psi_{jj}) \nonumber \\
        &=& \frac{2}{h} \cdot \trace\left(\boldsymbol \Omega \oplus
        \boldsymbol \Psi\right).
    \end{eqnarray}
    This completes the proof of the theorem.
\end{proof}

With certain normalization on $\boldsymbol X$, we can have
$\overline{\boldsymbol \mu} + \overline{\boldsymbol \upsilon} = 0$,
resulting in
\begin{equation}
    \tilde{\boldsymbol M} = \boldsymbol \mu \diamondplus
    \boldsymbol \upsilon.
\end{equation}
As the trace of a covariance matrix measures
the total variation, Theorem 1 implies that $\tilde{\boldsymbol X}$
follows a matrix-variate normal distribution with the same mean and
scaled covariance as the distribution of $\boldsymbol X$ in
Eq.~(\ref{eqn:mvn_1}). Given this conclusion and the process to
obtain $\tilde{\boldsymbol X}$ from $\boldsymbol X$, we propose our
Kronecker attention operators in the following section.

\begin{figure*}[t] \includegraphics[width=\textwidth]{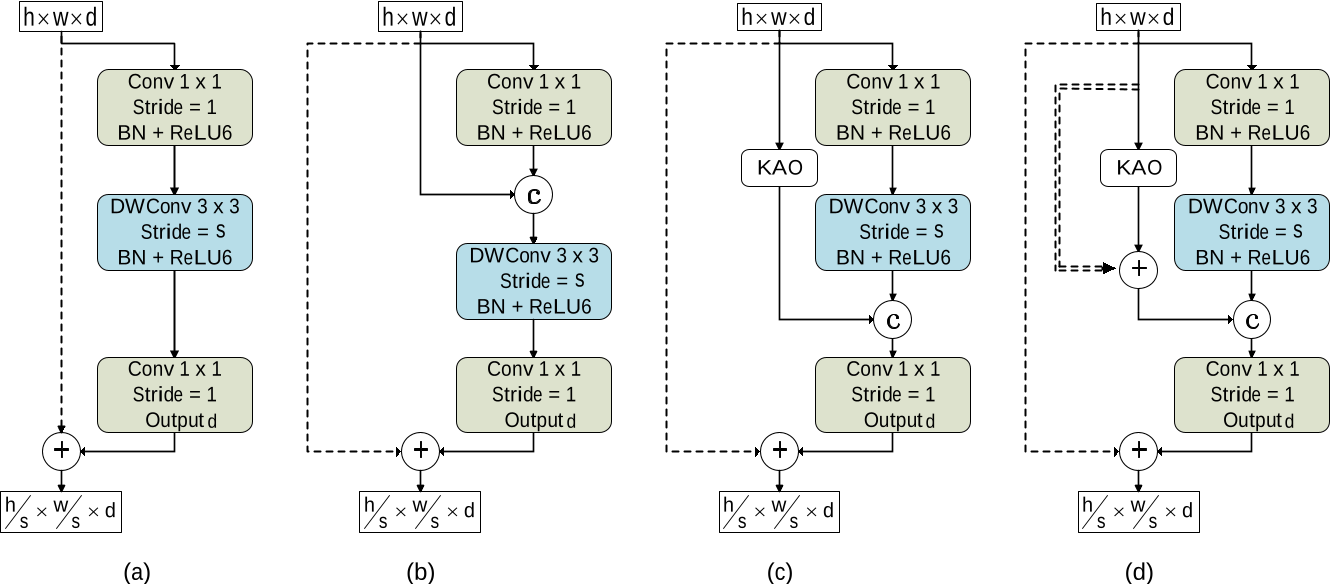}
    \Description[]{}
    \caption{Architectures
        of the BaseModule~(a), BaseSkipModule~(b), AttnModule~(c),
        and AttnSkipModule~(d) as described in
        Section~\ref{sec:module}. The skip connections indicated by
        single dashed paths are not used when $s > 1$ or $c \ne d$.
        Those indicated by double dashed paths are not used when $s >
            1$.} \label{fig:blocks}
\end{figure*}

\subsection{Kronecker Attention Operators}\label{sec:kals}
We describe the Kronecker attention operators (KAO) in the context of
self-attention on 2-D data, but they can be easily generalized to generic
attentions. In this case, the input to the $\ell$th layer is a third-order
tensor $\boldsymbol{\mathscr{X}}^{(\ell)} \in \mathbb{R}^{h\times w \times c}$.
Motivated by the theoretical results of Sections~\ref{sec:MCS}
and~\ref{sec:tecresults}, we propose to use horizontal and lateral average
matrices to represent original mode-3 unfolding without much information loss.
Based on Eq.~(\ref{eqn:r_mean}) and Eq.~(\ref{eqn:c_mean}), the horizontal
average matrix $\boldsymbol H$ and the lateral average matrix $\boldsymbol L$
are computed as
\begin{equation}
    \begin{aligned}
        \boldsymbol{H} = \frac{1}{h}
        \sum_{i=1}^{h}\boldsymbol{X}_{i::}^{(\ell)} \in \mathbb{R}^{w\times
            c}, \\
        \boldsymbol{L} = \frac{1}{w}
        \sum_{j=1}^{w}\boldsymbol{X}_{:j:}^{(\ell)} \in \mathbb{R}^{h\times
            c},
    \end{aligned}
\end{equation}
where $\boldsymbol{X}_{i::}^{(\ell)}$ and
$\boldsymbol{X}_{:j:}^{(\ell)}$ are the horizontal and lateral
slices~\cite{Tensor:Tamara} of tensor
$\boldsymbol{\mathscr{X}}^{(\ell)}$, respectively. We then form a
matrix $\boldsymbol{C}$ by juxtaposing $\boldsymbol{H}^T$ and
$\boldsymbol{L}^T$ as
\begin{equation}
    \boldsymbol{C} = [\boldsymbol H^T, \boldsymbol L^T] \in
    \mathbb{R}^{c \times (h+w)}.
\end{equation}

Based on the horizontal and lateral average matrices contained in
$\boldsymbol{C}$, we propose two Kronecker attention operators
(KAOs), \emph{i.e.}, KAO\textsubscript{{\tiny{$KV$}}} and
KAO\textsubscript{{\tiny{$QKV$}}}. In
KAO\textsubscript{{\tiny{$KV$}}} as shown in Figure~\ref{fig:atts}
(b), we use $\boldsymbol X_{(3)}^{(\ell)}$ as the query matrix and
$\boldsymbol C$ as the key and value matrices as
\begin{equation}\label{eq:att:kv}
    \boldsymbol O = \mbox{attn}(\boldsymbol X_{(3)}^{(\ell)},
    \boldsymbol{C}, \boldsymbol{C})\in\mathbb{R}^{c\times hw}.
\end{equation}
Note that the number of columns in $\boldsymbol O$ depends on the
number of query vectors. Thus, we obtain $hw$ output vectors from
the attention operation in Eq.~(\ref{eq:att:kv}). Similar to the
regular attention operator, $\boldsymbol O$ is folded back to a
third-order tensor $\boldsymbol{\mathscr{Y}}^{(\ell)} \in
    \mathbb{R}^{h \times w \times c}$ by considering the column vectors
in $\boldsymbol O$ as mode-3 fibers of
$\boldsymbol{\mathscr{Y}}^{(\ell)}$.
KAO\textsubscript{{\tiny{$KV$}}} uses
$\boldsymbol{\mathscr{Y}}^{(\ell)}$ as the output of layer $\ell$.

If $h = w$, the time and space complexities of
KAO\textsubscript{{\tiny{$KV$}}} are $O(hw\times c \times (h+w)) =
    O(h^3\times c)$ and $O(hw \times (h+w)) = O(h^3)$, respectively.
Compared to the original local operator on 2-D data,
KAO\textsubscript{{\tiny{$KV$}}} reduces time and space complexities
by a factor of $h$.

In order to reduce the time and space complexities further, we
propose another operator known as KAO\textsubscript{{\tiny{$QKV$}}}.
In KAO\textsubscript{{\tiny{$QKV$}}} as shown in
Figure~\ref{fig:atts}(c), we use $\boldsymbol C$ as the query, key,
and value matrices as
\begin{equation}\label{eq:kalv2}
    [\underbrace{\boldsymbol{\tilde H}}_{h},
        \underbrace{\boldsymbol{\tilde L}}_{w}] = \boldsymbol O =
    \mbox{attn}(\boldsymbol C, \boldsymbol C, \boldsymbol
    C)\in\mathbb{R}^{c\times (h+w)}.
\end{equation}
The final output tensor $\boldsymbol{\mathscr{Y}}^{(\ell)} \in
    \mathbb{R}^{h \times w \times c}$ is obtained as
\begin{equation}
    \boldsymbol{Y}^{(\ell)}_{::i} = \boldsymbol{\tilde H}^T_{i:}
    \diamondplus \boldsymbol{\tilde L}^T_{i:},
\end{equation}
where $\boldsymbol{\tilde H}_{i:}$ and $\boldsymbol{\tilde
        L}_{i:}$ are the $i$th rows of the corresponding matrices. That
is, the $i$th frontal slice of $\boldsymbol{\mathscr{Y}}^{(\ell)}$
is obtained by computing the outer sum of the $i$th rows of
$\boldsymbol{\tilde H}$ and $\boldsymbol{\tilde L}$.

If $h=w$, the time and space complexities of
KAO\textsubscript{{\tiny{$QKV$}}} are $O((h+w) \times c \times
    (h+w)) = O(h^2\times c)$ and $O((h+w) \times (h+w)) = O(h^2)$,
respectively. Thus, the time and space complexities have been
reduced by a factor of $h^2$ as compared to the original local
operator, and by a factor of $h$ as compared to
KAO\textsubscript{{\tiny{$KV$}}}.

Note that we do not consider
linear transformations in our description, but these
transformations can be applied to all three input matrices in
KAO\textsubscript{{\tiny{$KV$}}} and
KAO\textsubscript{{\tiny{$QKV$}}} as shown in Figure~\ref{fig:atts}.

\subsection{Kronecker Attention Modules and Networks}\label{sec:module}

\begin{table}[t]
    \centering \caption{Details of the KANets architecture. Each line describes a sequence of operators in the format of ``input size / operator name / expansion rate $\boldsymbol r$ / number of output channels $\boldsymbol c$ / number of operators in the sequence $\boldsymbol n$ / stride $\boldsymbol s$''. ``Conv2D'' denotes the regular 2D convolutional layer. ``AvgPool'' and ``FC'' denote the global average pooling layer and the fully-connected layer, respectively. All depth-wise convolutions use the kernel size of $3 \times 3$. For multiple operators in a sequence denoted in the same line, all operators produce $\boldsymbol c$ output channels. And the first operator applies the stride of $\boldsymbol s$ while the following operators applies the stride of 1. $k$ denotes the class number in the task.}
    \label{table:netbody}
    \begin{tabularx}{\columnwidth}{  l  l  YYYY }
        \hline
        \textbf{Input}  & \textbf{Operator} & $\boldsymbol r$ & $\boldsymbol c$ & $\boldsymbol n$ & $\boldsymbol s$ \\ \hline\hline
        224$^2\times$3  & Conv2D $3\times3$ & -               & 32              & 1               & 2               \\ \hline
        112$^2\times$32 & BaseSkipModule    & 1               & 16              & 1               & 1               \\ \hline
        112$^2\times$16 & BaseSkipModule    & 6               & 24              & 2               & 2               \\ \hline
        56$^2\times$24  & BaseSkipModule    & 6               & 32              & 2               & 2               \\ \hline
        28$^2\times$32  & AttnSkipModule    & 6               & 32              & 1               & 1               \\ \hline
        28$^2\times$32  & BaseSkipModule    & 6               & 64              & 1               & 2               \\ \hline
        14$^2\times$64  & AttnSkipModule    & 6               & 64              & 3               & 1               \\ \hline
        14$^2\times$64  & AttnSkipModule    & 6               & 96              & 3               & 1               \\ \hline
        14$^2\times$96  & BaseSkipModule    & 6               & 160             & 1               & 2               \\ \hline
        7$^2\times$160  & AttnSkipModule    & 6               & 160             & 2               & 1               \\ \hline
        7$^2\times$160  & AttnSkipModule    & 6               & 320             & 1               & 1               \\ \hline
        7$^2\times$320  & Conv2D $1\times1$ & -               & 1280            & 1               & 1               \\ \hline
        7$^2\times$1280 & AvgPool + FC      & -               & $k$             & 1               & -               \\ \hline
        \hline
    \end{tabularx}
\end{table}

Attention models have not been used in compact deep models to date,
primarily due to their high computational cost. Our efficient KAOs
make it possible to use attention operators in compact convolutional
neural networks (CNNs) like MobileNet~\cite{sandler2018mobilenetv2}.
In this section, we design a family of efficient Kronecker attention
modules based on MobileNetV2 that can be used in compact CNNs.

\textbf{BaseModule:} MobileNetV2~\cite{sandler2018mobilenetv2} is
mainly composed of bottleneck blocks with inverted residuals. Each
bottleneck block consists of three convolutional layers; those are,
$1\times1$ convolutional layer, $3\times3$ depth-wise convolutional
layer, and another $1\times1$ convolutional layer. Suppose the
expansion factor is $r$ and stride is $s$. Given input
$\boldsymbol{\mathscr{X}}^{(\ell)} \in \mathbb{R}^{h\times w \times
        c}$ for the $\ell$th block, the first $1\times1$ convolutional layer
outputs $rc$ feature maps $\boldsymbol{\mathscr{\tilde X}}^{(\ell)}
    \in \mathbb{R}^{h\times w \times rc}$. The depth-wise convolutional
layer uses a stride of $s$ and outputs $rc$ feature maps
$\boldsymbol{\mathscr{\bar X}}^{(\ell)} \in \mathbb{R}^{\frac{h}{s}
    \times \frac{w}{s} \times rc}$. The last $1\times1$ convolutional
layer produces $d$ feature maps $\boldsymbol{\mathscr{Y}}^{(\ell)}
    \in \mathbb{R}^{\frac{h}{s}\times \frac{w}{s} \times d}$. When $s=1$
and $c=d$, a skip connection is added between
$\boldsymbol{\mathscr{X}}^{(\ell)}$ and
$\boldsymbol{\mathscr{Y}}^{(\ell)}$. The BaseModule is illustrated
in Figure~\ref{fig:blocks}~(a).

\textbf{BaseSkipModule:} To facilitate feature reuse and gradient
back-propagation in deep models, we improve the BaseModule by adding
a skip connection. Given input $\boldsymbol{\mathscr{X}}^{(\ell)}$,
we use an expansion factor of $r-1$ for the first $1\times1$
convolutional layer, instead of $r$ as in BaseModule. We then
concatenate the output with the original input, resulting in
$\boldsymbol{\mathscr{\tilde X}}^{(\ell)} \in \mathbb{R}^{h\times w
        \times rc}$. The other parts of the BaseSkipModule are the same as
those of the BaseModule as illustrated in
Figure~\ref{fig:blocks}~(b). Compared to the BaseModule, the
BaseSkipModule reduces the number of parameters by $c \times c$ and
computational cost by $h \times w \times c$. It achieves better
feature reuse and gradient back-propagation.

\textbf{AttnModule:} We propose to add an attention operator into
the BaseModule to enable the capture of global features. We reduce
the expansion factor of the BaseModule by $1$ and add a new parallel
path with an attention operator that outputs $c$ feature maps.
Concretely, after the depth-wise convolutional layer, the original
path outputs $\boldsymbol{\mathscr{\bar X}}^{(\ell)}_{a} \in
    \mathbb{R}^{\frac{h}{s} \times \frac{w}{s} \times (r-1)c}$. The
attention operator, optionally followed by an average pooling of
stride $s$ if $s
    > 1$, produces $\boldsymbol{\mathscr{\bar X}}^{(\ell)}_{b} \in
    \mathbb{R}^{\frac{h}{s} \times \frac{w}{s} \times c}$. Concatenating
them gives $\boldsymbol{\mathscr{\bar X}}^{(\ell)} \in
    \mathbb{R}^{\frac{h}{s}\times \frac{w}{s} \times rc}$. The final
$1\times1$ convolutional layer remains the same. Within the
attention operator, we only apply the linear transformation on the
value matrix $\boldsymbol V$ to limit the number of parameters and
required computational resources. We denote this module as the
AttnModule as shown in Figure~\ref{fig:blocks}~(b). In this module,
the original path acts as locality-based feature extractors, while
the new parallel path with an attention operator computes global
features. This enables the module to incorporate both local and
global information. Note that we can use any attention operator in
this module, including the regular attention operator and our KAOs.

\textbf{AttnSkipModule:} We propose to add an additional skip
connection in the AttnModule, as shown in
Figure~\ref{fig:blocks}~(d). This skip connection can always be
added unless $s > 1$. The AttnSkipModule has the same amount of
parameters and computational cost as the AttnModule.

\section{Experimental Studies}

In this section, we evaluate our proposed operators and networks on
image classification and segmentation tasks. We first compare our
proposed KAOs with regular attention operators in terms of
computational cost and memory usage. Next, we design novel compact
CNNs known as Kronecker attention networks~(KANets) using our
proposed operators and modules. We compare KANets with other compact
CNNs on the ImageNet ILSVRC 2012 dataset~\cite{imagenet_cvpr09}.
Ablation studies are conducted to investigate how our KAOs benefit
the entire networks. We also perform experiments on the PASCAL 2012
dataset~\cite{everingham2010pascal} to show the effectiveness of our
KAOs on general application scenarios.

\begin{table*}[t]
    \caption{Comparisons between the regular attention operator, the regular attention operator with a pooling operation~\cite{wang2018non}, and our proposed KAO\textsubscript{{\tiny{$KV$}}} and KAO\textsubscript{{\tiny{$QKV$}}} in terms of the number of parameters, number of MAdd, memory usage, and CPU inference time on simulated data of different sizes. The input sizes are given in the format of ``batch size $\times$ spatial sizes $\times$ number of input channels''. ``Attn'' denotes the regular attention operator. ``Attn+Pool'' denotes the regular attention operator which employs a $2\times2$ pooling operation on $\boldsymbol K$ and $\boldsymbol V$ input matrices to reduce required computational resources.}
    \label{table:layers_cmp}
    \begin{tabularx}{\textwidth}{Y X  Y Y Y c Y Y}
        \hline
        \textbf{Input} & \textbf{Operator}                 & \textbf{MAdd}  & \textbf{Cost~Saving} & \textbf{Memory} & \textbf{Memory~Saving} & \textbf{Time}  & \textbf{Speedup}       \\ \hline\hline
        \multirow{4}{*}{$8\times14^2\times8$}
                       & Attn                              & 0.63m          & 0.00\%               & 5.2MB           & 0.00\%                 & 5.8ms          & 1.0$\times$            \\ 
                       & Attn+Pool                         & 0.16m          & 75.00\%              & 1.5MB           & 71.65\%                & 2.0ms          & 3.0$\times$            \\ 
                       & KAO\textsubscript{{\tiny{$KV$}}}  & 0.09m          & 85.71\%              & 0.9MB           & 82.03\%                & 1.7ms          & 3.5$\times$            \\ 
                       & KAO\textsubscript{{\tiny{$QKV$}}} & \textbf{0.01m} & \textbf{97.71\%}     & \textbf{0.3MB}  & \textbf{95.06\%}       & \textbf{0.8ms} & \textbf{6.8$\times$}   \\ \hline
        \multirow{4}{*}{$8\times28^2\times8$}
                       & Attn                              & 9.88m          & 0.00\%               & 79.9MB          & 0.00\%                 & 72.4ms         & 1.0$\times$            \\ 
                       & Attn+Pool                         & 2.47m          & 75.00\%              & 20.7MB          & 74.13\%                & 20.9ms         & 3.5$\times$            \\ 
                       & KAO\textsubscript{{\tiny{$KV$}}}  & 0.71m          & 92.86\%              & 6.5MB           & 91.88\%                & 7.1ms          & 10.1$\times$           \\ 
                       & KAO\textsubscript{{\tiny{$QKV$}}} & \textbf{0.05m} & \textbf{99.46\%}     & \textbf{0.9MB}  & \textbf{98.85\%}       & \textbf{1.7ms} & \textbf{40.9$\times$}  \\ \hline
        \multirow{4}{*}{$8\times56^2\times8$}
                       & Attn                              & 157.55m        & 0.00\%               & 1,262.6MB       & 0.00\%                 & 1,541.1ms      & 1.0$\times$            \\ 
                       & Attn+Pool                         & 39.39m         & 75.00\%              & 318.7MB         & 74.76\%                & 396.9ms        & 3.9$\times$            \\ 
                       & KAO\textsubscript{{\tiny{$KV$}}}  & 5.62m          & 96.43\%              & 48.2MB          & 96.18\%                & 49.6ms         & 31.1$\times$           \\ 
                       & KAO\textsubscript{{\tiny{$QKV$}}} & \textbf{0.21m} & \textbf{99.87\%}     & \textbf{3.4MB}  & \textbf{99.73\%}       & \textbf{5.1ms} & \textbf{305.8$\times$} \\ \hline
        \hline
    \end{tabularx}
\end{table*}

\subsection{Experimental Setup}\label{sec:setup}

In this section, we describe the experimental setups for both
image classification tasks and image segmentation tasks.

\textbf{Experimental Setup for Image Classification}
As a common practice on this dataset, we use the same data augmentation scheme
in~\citet{he2016deep}. Specifically, during training, we scale each image to
$256\times256$ and then randomly crop a $224\times224$ patch. During inference,
the center-cropped patches are used. We train our KANets using the same settings
as MobileNetV2~\cite{sandler2018mobilenetv2} with minor changes. We perform
batch normalization~\cite{ioffe2015batch} on the coefficient matrices in KAOs to
stabilize the training. All trainable parameters are initialized with the Xavier
initialization~\cite{glorot2010understanding}. We use the standard stochastic
gradient descent optimizer with a momentum of 0.9~\cite{sutskever2013importance}
to train models for 150 epochs in total. The initial learning rate is 0.1 and it
decays by 0.1 at the $80$th, $105$th, and $120$th epoch.
Dropout~\cite{srivastava2014dropout} with a keep rate of $0.8$ is applied after
the global average pooling layer. We use 8 TITAN Xp GPUs and a batch size of
$512$ for training, which takes about $1.5$ days. Since labels of the test
dataset are not available, we train our networks on training dataset and report
accuracies on the validation dataset.

\textbf{Experimental Setup for Image Segmentation}
We train all the models with randomly cropped patches of size $321 \times 321$
and a batch size of 8. Data augmentation by randomly scaling the inputs for
training is employed. We adopt the ``poly'' learning rate
policy~\cite{liu2015parsenet} with $power=0.9$, and set the initial learning
rate to 0.00025. Following DeepLabV2, we use the ResNet-101 model  pre-trained
on ImageNet~\cite{imagenet_cvpr09} and MS-COCO~\cite{lin2014microsoft} for
initialization. The models are then trained for 25,000 iterations with a
momentum of 0.9 and a weight decay of 0.0005. We perform no post-processing such
as conditional random fields and do not use multi-scale inputs due to limited
GPU memory. All the models are trained on the training set and evaluated on the
validation set.

\subsection{Comparison of Computational Efficiency}\label{sec:comp_att}
According to the theoretical analysis in Section~\ref{sec:kals}, our
KAOs have efficiency advantages over regular attention operators on
high-order data, especially for inputs with large spatial sizes. We
conduct simulated experiments to evaluate the theoretical results.
To reduce the influence of external factors, we build networks
composed of a single attention operator, and apply the TensorFlow
profile tool~\cite{abadi2016tensorflow} to report the
multiply-adds~(MAdd), required memory, and time consumed on 2-D
simulated data. For the simulated input data, we set the batch size
and number of channels both to 8, and test three spatial sizes;
those are, $56 \times 56$, $28 \times 28$, and $14 \times 14$. The
number of output channels is also set to 8.

Table~\ref{table:layers_cmp} summarizes the comparison results. On
simulated data of spatial sizes $56\times56$, our
KAO\textsubscript{{\tiny{$KV$}}} and
KAO\textsubscript{{\tiny{$QKV$}}} achieve 31.1 and 305.8 times
speedup, and 96.18\% and 99.73\% memory saving compared to the
regular attention operator, respectively. Our proposed KAOs show
significant improvements over regular attention operators in terms
of computational resources, which is consistent with the theoretical
analysis. In particular, the amount of improvement increases as the
spatial sizes increase. These results show that the proposed KAOs
are efficient attention operators on high-dimensional and high-order
data.

\begin{table}[t]
    \caption{Comparisons between KANets and other CNNs in terms of the top-1 accuracy on the ImageNet validation set, the number of total parameters, and MAdd. We use KANet\textsubscript{{\tiny{$KV$}}} and KANet\textsubscript{{\tiny{$QKV$}}} to denote KANets using KAO\textsubscript{{\tiny{$KV$}}} and KAO\textsubscript{{\tiny{$QKV$}}}, respectively.}
    \label{table:cmp_models}
    \begin{tabularx}{\columnwidth}{  l   Y  Y  Y }
        \hline
        \textbf{Model}                                                   & \textbf{Top-1} & \textbf{Params} & \textbf{MAdd} \\ \hline\hline
        GoogleNet                                                        & 0.698          & 6.8m            & 1550m         \\ \hline
        VGG16                                                            & 0.715          & 128m            & 15300m        \\ \hline
        AlexNet                                                          & 0.572          & 60m             & 720m          \\ \hline\hline
        SqueezeNet                                                       & 0.575          & 1.3m            & 833m          \\ \hline
        MobileNetV1                                                      & 0.706          & 4.2m            & 569m          \\ \hline
        ShuffleNet 1.5x                                                  & 0.715          & 3.4m            & 292m          \\ \hline
        ChannelNet-v1                                                    & 0.705          & 3.7m            & 407m          \\ \hline
        MobileNetV2                                                      & 0.720          & 3.47m           & 300m          \\ \hline
        \textbf{KANet\textsubscript{{\tiny{$\boldsymbol{KV}$}}}} (ours)  & \textbf{0.729} & 3.44m           & 288m          \\ \hline
        \textbf{KANet\textsubscript{{\tiny{$\boldsymbol{QKV}$}}}} (ours) & 0.728          & 3.44m           & \textbf{281m} \\ \hline
        \hline
    \end{tabularx}
\end{table}

\subsection{Results on Image Classification}

With the high efficiency of our KAOs, we have proposed several
efficient Kronecker attention modules for compact CNNs in
Section~\ref{sec:module}. To further show the effectiveness of KAOs
and the modules, we build novel compact CNNs known as Kronecker
attention networks~(KANets). Following the practices
in~\cite{wang2018non}, we apply these modules on inputs of spatial
sizes $28\times28$, $14\times14$, and $7\times7$. The detailed
network architecture is described in Table~\ref{table:netbody} in
the Section~\ref{sec:setup}.

We compare KANets with other CNNs on the ImageNet ILSVRC 2012 image
classification dataset, which serves as the benchmark for compact
CNNs~\cite{howard2017mobilenets,zhang2017shufflenet,gao2018channelnets,sandler2018mobilenetv2}.
The dataset contains 1.2 million training, 50 thousand validation,
and 50 thousand testing images. Each image is labeled with one of
1,000 classes. Details of the experimental setups are provided in
the Section~\ref{sec:setup}.

The comparison results between our KANets and other CNNs in terms of
the top-1 accuracy, number of parameters, and MAdd are reported in
Table~\ref{table:cmp_models}.
SqueezeNet~\cite{iandola2016squeezenet} has the least number of
parameters, but uses the most MAdd and does not obtain competitive
performance as compared to other compact CNNs. Among compact CNNs,
MobileNetV2~\cite{sandler2018mobilenetv2} is the previous
state-of-the-art model, which achieves the best trade-off between
effectiveness and efficiency. According to the results, our KANets
significantly outperform MobileNetV2 with 0.03 million fewer
parameters. Specifically, our KANet\textsubscript{{\tiny{$KV$}}} and
KANet\textsubscript{{\tiny{$QKV$}}} outperform MobileNetV2 by
margins of 0.9\% and 0.8\%, respectively. More importantly, our
KANets has the least computational cost. These results demonstrate
the effectiveness and efficiency of our proposed KAOs.

The performance of KANets indicates that our proposed methods are
promising, since we only make small modifications to the
architecture of MobileNetV2 to include KAOs. Compared to modules
with the regular convolutional layers only, our proposed modules
with KAOs achieve better performance without using excessive
computational resources. Thus, our methods can be used widely for
designing compact deep models. Our KAOs successfully address the
practical challenge of applying regular attention operators on
high-order data. In the next experiments, we show that our proposed
KAOs are as effective as regular attention operators.

\begin{table}[t]
    \caption{Comparisons between KANets with regular attention operators~(denoted as AttnNet), KANets with regular attention operators with a pooling operation~(denoted as AttnNet+Pool) and KANets with KAOs in terms of the top-1 accuracy on the ImageNet validation set, the number of total parameters, and MAdd.}
    \label{table:atts_cmp}
    \begin{tabularx}{\columnwidth}{ l Y Y Y Y }
        \hline
        \textbf{Model}                      & \textbf{Top-1} & \textbf{Params} & \textbf{MAdd} \\ \hline\hline
        AttnNet                             & 0.730          & 3.44m           & 365m          \\ \hline
        AttnNet+Pool                        & 0.729          & 3.44m           & 300m          \\ \hline
        KANet\textsubscript{{\tiny{$KV$}}}  & 0.729          & 3.44m           & 288m          \\ \hline
        KANet\textsubscript{{\tiny{$QKV$}}} & 0.728          & 3.44m           & \textbf{281m} \\ \hline
        \hline
    \end{tabularx}
\end{table}

\subsection{Comparison with Regular Attention Operators}
We perform experiments to compare our proposed KAOs with regular
attention operators. We consider the regular attention operator and
the one with a pooling operation in~\cite{wang2018non}. For the
attention operator with pooling operation, the spatial sizes of the
key matrix $\boldsymbol K$ and value matrix $\boldsymbol V$ are
reduced by $2 \times 2$ pooling operations to save computation cost.
To compare these operators in fair settings, we replace all KAOs in
KANets with regular attention operators and regular attention
operators with a pooling operation, denoted as AttnNet and
AttnNet+Pool, respectively.

The comparison results are summarized in Table~\ref{table:atts_cmp}.
Note that all these models have the same number of parameters. We
can see that KANet\textsubscript{{\tiny{$KV$}}} and
KANet\textsubscript{{\tiny{$QKV$}}} achieve similar performance as
AttnNet and AttnNet+Pool with dramatic reductions of
computational cost. The results indicate that our proposed KAOs are
as effective as regular attention operators while being much more
efficient. In addition, our KAOs are better than regular attention
operators that uses a pooling operation to increase efficiency
in~\cite{wang2018non}.

\subsection{Ablation Studies}

To show how our KAOs benefit entire networks in different settings,
we conduct ablation studies on MobileNetV2 and
KANet\textsubscript{{\tiny{$KV$}}}. For MobileNetV2, we replace
BaseModules with AttnModules as described in
Section~\ref{sec:module}, resulting in a new model denoted as
MobileNetV2+KAO. On the contrary, based on
KANet\textsubscript{{\tiny{$KV$}}}, we replace all AttnSkipModules
by BaseModules. The resulting model is denoted as
KANet~w/o~KAO.

Table~\ref{table:abla} reports the comparison results. By employing
KAO\textsubscript{{\tiny{$KV$}}}, MobileNetV2+KAO gains a
performance boost of 0.6\% with fewer parameters than MobileNetV2.
On the other hand, KANet\textsubscript{{\tiny{$KV$}}} outperforms
KANet~w/o~KAO by a margin of 0.8\%, while
KANet~w/o~KAO has more parameters than
KANet\textsubscript{{\tiny{$KV$}}}.
KANet\textsubscript{{\tiny{$KV$}}} achieves the best performance
while costing the least computational resources. The results
indicate that our proposed KAOs are effective and efficient, which
is independent of specific network architectures.

\begin{table}[t]
    \caption{Comparisons between MobileNetV2, MobileNetV2 with
        KAOs\textsubscript{{\tiny{$KV$}}} (denoted as
        MobileNetV2+KAO\textsubscript{{\tiny{$KV$}}}),
        KANet\textsubscript{{\tiny{$KV$}}}, and
        KANet\textsubscript{{\tiny{$KV$}}} without
        KAO\textsubscript{{\tiny{$KV$}}} (denoted as KANet~w/o~KAO) in terms
        of the top-1 accuracy on the ImageNet validation set, the number of
        total parameters, and MAdd.}
    \label{table:abla}
    \begin{tabularx}{\columnwidth}{  l   Y  Y  Y }
        \hline
        \textbf{Model}                     & \textbf{Top-1} & \textbf{Params} & \textbf{MAdd} \\ \hline\hline
        MobileNetV2                        & 0.720          & 3.47m           & 300m          \\ \hline
        MobileNetV2+KAO                    & 0.726          & 3.46m           & 298m          \\ \hline
        KANet\textsubscript{{\tiny{$KV$}}} & \textbf{0.729} & \textbf{3.44m}  & \textbf{288m} \\ \hline
        KANet~w/o~KAO                      & 0.721          & 3.46m           & 298m          \\ \hline
        \hline
    \end{tabularx}
\end{table}

\subsection{Results on Image Segmentation}

\begin{table}[t] \caption{Comparisons of DeepLabV2, DeepLabV2
        with the regular attention operator (DeepLabV2+Attn), DeepLabV2
        with
        our KAO\textsubscript{{\tiny{$KV$}}} (DeepLabV2+KAO\textsubscript{{\tiny{$KV$}}}),
        and DeepLabV2 with our KAO\textsubscript{{\tiny{$QKV$}}}
        (DeepLabV2+KAO\textsubscript{{\tiny{$QKV$}}})
        in terms of the pixel-wise accuracy, and mean IOU on the PASCAL
        VOC 2012 validation dataset.}
    \label{table:seg_cmp}
    \begin{tabularx}{\columnwidth}{  l  Y  Y  Y }
        \hline
        \textbf{Model}                              & \textbf{Accuracy} & \textbf{Mean IOU} \\ \hline\hline
        DeepLabV2                                   & 0.944             & 75.1              \\ \hline
        DeepLabV2+Attn                              & 0.947             & 76.3              \\ \hline
        DeepLabV2+KAO\textsubscript{{\tiny{$KV$}}}  & 0.946             & 75.9              \\ \hline
        DeepLabV2+KAO\textsubscript{{\tiny{$QKV$}}} & 0.946             & 75.8              \\ \hline
        \hline
    \end{tabularx}
\end{table}

In order to show the efficiency and effectiveness of our KAOs in
broader application scenarios, we perform additional experiments on
image segmentation tasks using the PASCAL 2012
dataset~\cite{everingham2010pascal}. With the extra annotations
provided by~\cite{hariharan2011semantic}, the augmented dataset
contains 10,582 training, 1,449 validation, and 1,456 testing
images. Each pixel of the images is labeled by one of 21 classes
with 20 foreground classes and 1 background class.

We re-implement the DeepLabV2 model~\cite{chen2018deeplab} as our
baseline. Following~\cite{wang2018smoothed}, using attention
operators as the output layer, instead of atrous spatial pyramid
pooling (ASPP), results in a significant performance improvement. In
our experiments, we replace ASPP with the regular attention operator
and our proposed KAOs, respectively, and compare the results. For
all attention operators, linear transformations are applied on
$\boldsymbol Q$, $\boldsymbol K$, and $\boldsymbol V$. Details of
the experimental setups are provided in the Section~\ref{sec:setup}.

Table~\ref{table:seg_cmp} shows the evaluation results in terms of
pixel accuracy and mean intersection over union~(IoU) on the PASCAL
VOC 2012 validation set. Clearly, models with attention operators
outperform the baseline model with ASPP. Compared with the regular
attention operator, KAOs result in similar pixel-wise accuracy but
slightly lower mean IoU. From the pixel-wise accuracy, results
indicate that KAOs are as effective as the regular attention
operator. The decrease in mean IoU may be caused by the strong
structural assumption behind KAOs. Overall, the experimental results
demonstrate the efficiency and effectiveness of our KAOs in broader
application scenarios.

\section{Conclusions}

In this work, we propose Kronecker attention operators to address
the practical challenge of applying attention operators on
high-order data. We investigate the problem from a probabilistic
perspective and use matrix-variate normal distributions with
Kronecker covariance structure. Experimental results show that our
KAOs reduce the amount of required computational resources by a
factor of hundreds, with larger factors for higher-dimensional and
higher-order data. We employ KAOs to design a family of efficient
modules, leading to our KANets. KANets significantly outperform the
previous state-of-the-art compact models on image classification
tasks, with fewer parameters and less computational cost.
Additionally, we perform experiments on the image segmentation task
to show the effectiveness of our KAOs on general application
scenarios.

\begin{acks}
    This work was supported in part by National Science
    Foundation grants IIS-1908220 and DBI-1922969.
\end{acks}

\bibliographystyle{stys/ACM-Reference-Format}
\balance
\bibliography{stys/nips}

\end{document}